\documentclass{article} 
\usepackage[final]{colm2026_conference}

\usepackage[T2A,T1]{fontenc}    
\usepackage[utf8]{inputenc}  

\ifdefined\pdfmapline
  \pdfmapline{+larm0900 SFRM0900 "T2AEncoding ReEncodeFont" <cm-super-t2a.enc <sfrm0900.pfb}
  \pdfmapline{+larm1000 SFRM1000 "T2AEncoding ReEncodeFont" <cm-super-t2a.enc <sfrm1000.pfb}
\fi
\usepackage{microtype}
\usepackage{hyperref}
\usepackage{url}
\usepackage{doi}          
\usepackage{booktabs}
\usepackage{longtable}
\usepackage{multirow}
\usepackage{amsmath}
\usepackage{graphicx}
\usepackage{cleveref}
\usepackage{caption}
\usepackage{placeins}
\usepackage[russian,english]{babel} 
\usepackage{tcolorbox}

\newtcolorbox{examplebox}[2][]{
    colback=gray!10,
    colframe=gray!80,
    coltitle=white,
    fontupper=\footnotesize,
    fonttitle=\bfseries,
    title=#2,                      
    #1
}

\usepackage{lineno}

\definecolor{darkblue}{rgb}{0, 0, 0.5}
\hypersetup{colorlinks=true, citecolor=darkblue, linkcolor=darkblue, urlcolor=darkblue}

\title{We Hebben Een Serieus Translatie: Modeling \\ Intercomprehension as Probabilistic Inference}

\author{Thomas Hikaru Clark, Edward Gibson, \& Roger Levy \\
Department of Brain and Cognitive Sciences\\
MIT\\
Cambridge, MA 02139, USA \\
\texttt{\{thclark,egibson,rplevy\}@mit.edu}
}

\begin{document}

\ifcolmsubmission
\linenumbers
\fi

\maketitle

\begin{abstract}
    Intercomprehension refers to partial intelligibility of an unfamiliar language (L2) by a speaker of a related language (L1). How is this zero-shot cross-language comprehension possible? In this work, we extend past work on algorithmic models of noisy-channel inference to model intercomprehension in a Bayesian framework. The model uses an LM in L1 only for scoring latent hypotheses about the translations of observed L2 utterances, and a general-purpose noise model to infer a mapping between L2 and L1 words based on either form-based similarity or symbolic rules. We then conduct a human behavioral experiment, eliciting inferences for utterances in Dutch, Italian, and Ukrainian from speakers of English, Spanish, and Russian, respectively. 
    Our full model shows a closer alignment to the distribution of human intercomprehension performance than ablations, and also compares favorably to zero-shot prompting of much larger models.
    These results provide a cognitively plausible computational model of intercomprehension, and highlight the flexible inferences made by comprehenders under wide uncertainty in real-world cross-language scenarios. We share our code publicly.\footnote{Github: \url{https://github.com/thomashikaru/intercomprehension-colm-2026}} 
\end{abstract}

\section{Introduction}

Natural languages do not exist in a vacuum. As a result of the process of language evolution, languages exist in relation to each other in a hierarchy of linguistic proximity. Related languages often have some degree of \textbf{mutual intelligibility} or \textbf{intercomprehension} with each other, meaning that speakers of a language L1 can understand a non-trivial amount of a related language L2, without any formal training or exposure to L2 \citep{gooskensMutualIntelligibilityClosely2018, bonvinoObservingStrategiesIntercomprehension2018, carloIntercomprehensionStrengthsOpportunities2021, golubovicMutualIntelligibilityWest2015, beijeringPredictingIntelligibilityPerceived2008}. Given that this process can be asymmetric, rather than mutual, we will use the term \textbf{intercomprehension} as a general term for partial understanding of an unfamiliar or unstudied but related language. 

From a cognitive science perspective, this ability is a striking phenomenon, reflecting an ability to process and make rich inferences about out-of-distribution observations with large amount of uncertainty \citep{tenenbaumGeneralizationSimilarityandBayesian2001, ruleChildHacker2020, ellisUnsupervisedLearningProgram2015, tenenbaumHowGrowMind2011, xuWordLearningBayesian2007, zhi-xuanOnlineBayesianGoal2020}.
From a practical perspective, understanding the process by which the human mind performs this type of inference has ramifications for language pedagogy and inter-cultural communication and collaboration \citep{doyeMethodologicalFrameworkTeaching2004}.
In natural language processing, the problem of translating unknown languages is known as \textit{decipherment}, and has applications for archaeological and historical research on untranslated documents in extinct languages \citep{raviDecipheringForeignLanguage2011, luoNeuralDeciphermentMinimumCost2019, naimFeatureBasedDeciphermentMachine2018, luoDecipheringUndersegmentedAncient2021}. 

To provide a motivating example, the Dutch phrase \textit{We hebben een serieus probleem} went viral for being intelligible to English speakers with no prior Dutch exposure \footnote{\url{https://knowyourmeme.com/memes/we-hebben-een-serieus-probleem}}. This simple sentence has identical word order to its corresponding English translation, and contains cognates and/or loanwords which have surface-level similarities to their English counterparts. While it may seem obvious that an English speaker would be able to infer the meaning of this sentence, this example poses some challenges for a satisfying cognitive account, especially at an algorithmic level of explanation \citep{marrVisionComputationalInvestigation1982}. For example, the Dutch word \textit{een} in fact bears little surface-level similarity to its English counterpart \textit{a}, yet comprehenders appear able to infer its meaning ``from context''. It remains underspecified how exactly a comprehender performs this inference using context. 
Similarly, the word \textit{serieus} is also similar to other English words, such as \textit{series}, not just the correct translation of \textit{serious}. Thus, it appears that comprehenders are able to rapidly integrate multiple sources of information to produce an interpretation that aligns with probabilistic inference.

In this paper, we extend the \textbf{noisy-channel language processing} framework \citep{gibsonRationalIntegrationNoisy2013, levyNoisyChannelModelHuman2008, chenEffectContextNoisychannel2023} from psycholinguistics and cognitive science to the problem of intercomprehension. 
A basic premise of this framework is that humans make use of both a linguistic \textbf{prior} (expectations over which messages are likely to be sent) as well as a \textbf{likelihood} rooted in an implicit or explicit error model which captures the relative probabilities of different kinds of transformations that could ``corrupt'' or alter the intended message.
Recent work in this paradigm has offered candidate algorithmic models for simulating the fine-grained, incremental behavior of noisy-channel comprehenders under varying assumptions \citep{clarkModelApproximateIncremental2025, clarkResourceRationalNoisyChannelLanguage2025}, yet to our knowledge this framework has not been applied to the more general types of ``noise'' that comprehenders deal with in the context of intercomprehension.

Past work has modeled mutual intelligibility computationally by using Levenshtein distance \citep{gooskensPerceptiveEvaluationLevenshtein2004, beijeringPredictingIntelligibilityPerceived2008}. Yet edit distance alone only captures the component of intelligibility that relies upon surface form similarity, and leaves out an important variable: prior expectations about intended messages. 
It remains unclear if human comprehension of sentences in an unfamiliar language will reflect naive, surface-form-based inferences about its meaning, or integrated rational inference. 
Furthermore, an open question is which specific algorithmic components of inference most closely predict the pattern of successes and failures of humans at intercomprehension. 
\citet{niederComputationalModelAssessment2024} propose a model based on linear discriminative learning, but it operates on a dataset of cognates (rather than generating translations of arbitrary sentences). Meanwhile, \citet{luoNeuralDeciphermentMinimumCost2019} propose a decipherment model for ancient lost languages, but it is based on handcrafted patterns of historical sound change (rather than being a cognitive model of a naive zero-shot comprehender), and again is evaluated based on aligning cognates (percentage of cognates identified) rather than on translating full sentences. 
In this work, our unique contribution is an implemented cognitive model of intercomprehension that operates on arbitrary natural sentences, that integrates both form-based similarity and linguistic priors to form inferences. 

\section{Model: Tempered SMC for Zero-Shot Decipherment}

\subsection{Generative Model}

We set up a generative model, which expresses the process by which sentences from one language L1 may be transformed into sentences from another language L2. We assume that a sentence $z$ in L1 is sampled from a language model, which is simply a probability distribution over strings $p(z)$. 
In practice, we use simple, autoregressive language models from Goldfish \citep{changGoldfishMonolingualLanguage2026}, which provide monolingual models across 350 languages that are matches for parameter count and training data size. 
Since the purpose of the LM is only to capture the probability distribution over strings, not to perform reasoning or inference, we use a language model without Chain-of-Thought \citep{weiChainofThoughtPromptingElicits2023}, instruction-tuning, or extensive multilingual training data.

Given the L1 sentence $z$, the generative model applies transformations at the word level to yield a sentence $y$ in L2. 
The generative model reflects an L1 comprehender's intuitive theory of language transformations to an unknown but related language, thus the error model should be extremely general rather than reflecting any specific knowledge of the transformations that occur between a given pair of related languages.
For each word in $z$, the generative model randomly samples whether to apply a \textbf{string edit} transformation or a \textbf{memorized substitution} transformation. 

For a given word $z_t$, the distribution over string edit transformations is defined implicitly by the relationship $\log p(y_t \mid z_t) \propto - \text{editDistance}(y_t, z_t)$. Thus, larger string edits are less probable than smaller string edits, with each additional edit decreasing the probability exponentially. This reflects a naive intuitive theory of string transformations where edits are independent, random processes. 
We consider two distinct types of edit distance, one which instantiates an \textbf{orthographic likelihood} via string-based edit distance, and one which instantiates a \textbf{phonetic likelihood} by first converting L2 words to an ARPABET representation using the \verb|g2p_en| word pronunciation module in Python\footnote{\url{https://pypi.org/project/g2p-en/}, Version 2.1.0}, which predicts pronunciations for out-of-vocabulary words using a neural network, and computing edit distances in this space to ARPABET representations of items in an L1 wordlist. The final $\text{editDistance}()$
function uses an equally weighted average of orthographic and phonetic edit distance.

The memorized substitution transformation is a component of the generative model which reflects the idea that not all words have a cognate, even in highly related languages. Therefore, certain mappings between words in L1 and L2 simply need to be memorized. The generative model contains a \textbf{rule library} $\mathcal{L}$ which stores rewrite rules of the form $a \rightarrow b$. The probability distribution over rule libraries is defined implicitly by the relationship $\log p(\mathcal{L}) \propto \text{charCount}(\mathcal{L})$, enforcing a \textbf{minimum description length} prior: rule libraries that are simpler and more concise are more likely \textit{a priori} \citep{rissanenModelingShortestData1978, grunwaldMinimumDescriptionLength2007}. This prior additionally has the consequence that even if the same rule is used multiple times, it is only counted once in $\mathcal{L}$, implicitly rewarding rules which can explain multiple observations. 
The full generative model specification can be found in \Cref{app:model-spec}.

\subsection{SMC Inference}

Sequential Monte Carlo (SMC) is an inference algorithm for sampling from distributions that are challenging to sample from exactly, but where decomposing the problem into a sequence of sub-problems is beneficial for inference \citep{naessethElementsSequentialMonte2024,doucetSequentialMonteCarlo2001}. 
In simple terms, SMC involves maintaining a set of weighted \textbf{particles} which each correspond to a hypothesis about the value of latent random variables, and updating the particle weights sequentially while iterating through subproblems.
In many cases, SMC is applied to inference problems that are sequential in time, e.g. inferences about random variables that are conditioned on a temporal sequence of observations. This is the approach employed by \citet{clarkModelApproximateIncremental2025} to model psycholinguistic features of noisy-channel inference: possibly noisy words in a sentence are observed sequentially over time, while SMC updates a set of particles which correspond to a distribution over latent intended messages. 

For modeling intercomprehension, instead of modeling sequential observations, we model a sequence of inference targets where linguistic priors take on an increasingly prominent role. Given an observed L2 sentence $y$, we approximate the distribution over L1 translations of that sentence $z$. Thus, the target posterior is: $\pi_{\text{post}}(z \mid y) \propto p(y \mid z) p(z)$.
We set up a sequence of target distributions by introducing a parameter $\beta$ which acts as a tempering exponent on the prior, gradually increasing the prior's contribution \citep{nealAnnealedImportanceSampling1998, delmoralSequentialMonteCarlo2006}. This reflects the intuition that inference proceeds from solving an easier, more naive problem, i.e. identifying surface-level similarities between L1 and L2, to a more challenging and higher-level problem, i.e. combining likelihood and linguistic prior to identify strong candidate translations: $\pi_{\beta} \propto p(y \mid z) p(z)^\beta, \hspace{3mm} \beta \in [0, 1]$. 
$\pi_{\beta}$ now represents a family of targets. For $\beta = 0$, the language model prior does not contribute, and thus we initialize particles solely based on string edit distance, i.e. drawing from $\pi_0(z \mid y)$ by computing edit distances between observations $y_0 \dots y_T$ and words in an L1 wordlist. 
As $\beta$ is increased, the target distribution approaches the true posterior; at each step, particle weights are updated. Resampling occurs if the effective sample size (ESS) drops below a threshold. See \Cref{app:inference-details} for inference details. 

\subsection{MCMC Rejuvenation Moves}

\textbf{Rejuvenation} is performed after resampling to reintroduce diversity into the particles, without changing particle weights \citep{gilksFollowingMovingTarget2001,delmoralSequentialMonteCarlo2006}. Several specific MCMC kernels are used to apply rejuvenation. 
The \textbf{word substitution proposal} takes an observed word $y_t$ and simply samples a different word as the latent translation $z_t$ from the distribution over words in a L1 wordlist, with log probability inversely proportional to the edit distance with $y_t$.
The \textbf{common-word rule library proposal} takes an observed word $y_t$ and proposes a new library rule that it maps to an L1 function word $z_t$. 
The \textbf{contextual rule library proposal} takes an observed word $y_t$ in a latent context $z_{<t}$ and proposes a new library rule that maps it to an L1 word $z_t$ which is sampled from the L1 language model $p(\cdot \mid z_{<t})$. 
During each inference step, all three proposals are applied to each word in the observed sentence $y$ with the following exception: since the \textbf{contextual rule library proposal} critically depends on having a coherent sentential context in order to predict a missing word, it is only applied after a burn-in period of 2 iterations. 



\subsection{Example}

\begin{figure}[htb]
    \centering
    \includegraphics[width=\linewidth]{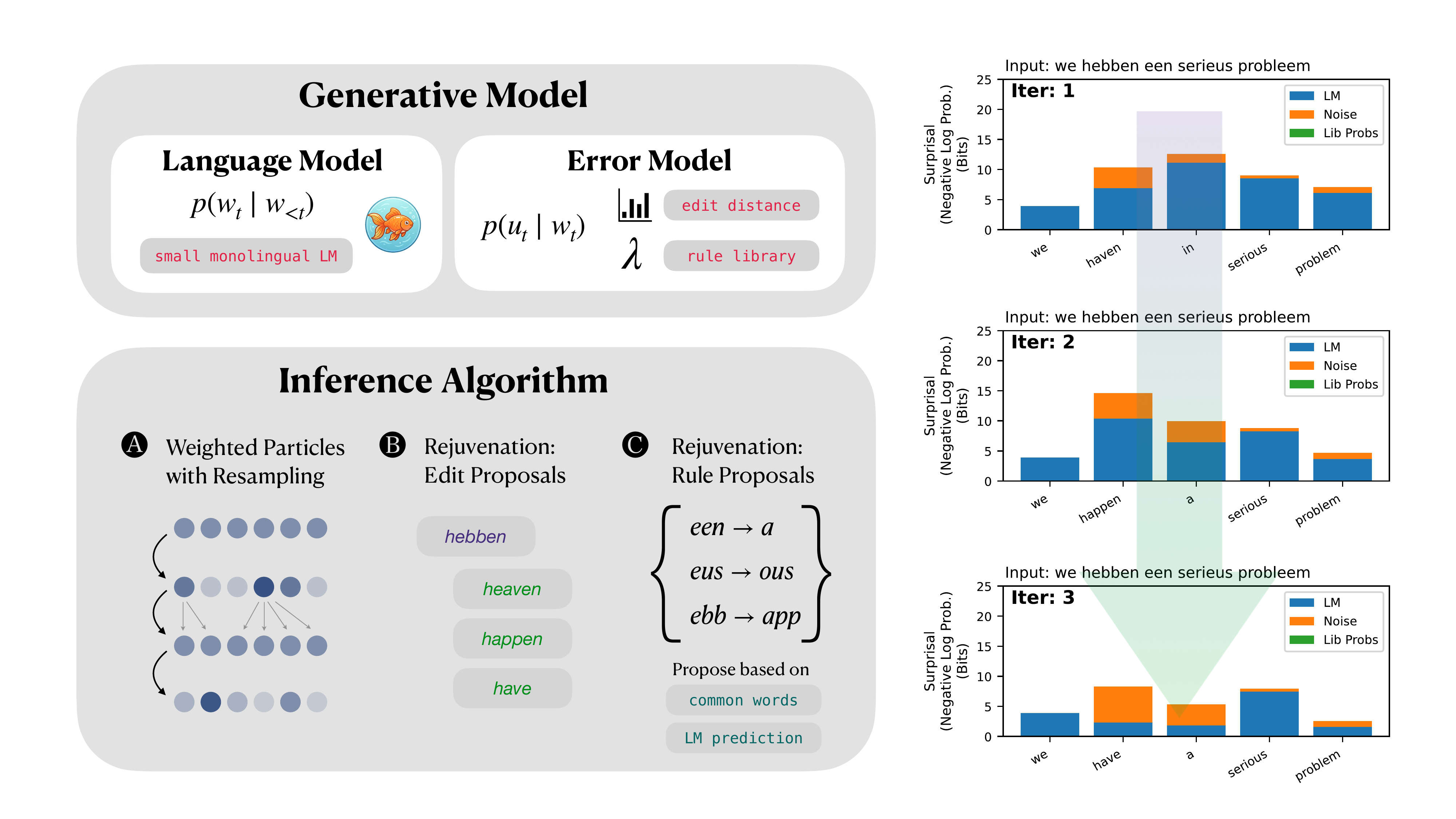}
    \caption{Model overview with example. \textbf{Generative model:} decomposed into a language model (prior) and noise model (likelihood). \textbf{Inference algorithm:} key components include SMC inference using weighted particles with resampling, and rejuvenation proposals that propose different word edits or adding/removing library rules. \textbf{Example:} Per-observation surprisal according to one particle over 3 successive iterations. Later stages of inference correspond to higher $\beta$, i.e. greater contribution of LM prior.}
    \label{fig:model-overview}
\end{figure}

We consider the progression of the inference process for an example Dutch sentence: \textit{We hebben een serieus probleem}. \Cref{fig:model-overview} visualizes this using a barplot showing the negative log probability, or \textbf{surprisal}, assigned by the model to each word in the observed Dutch sentence, decomposed into LM surprisal (derived from language model log-probabilities), library rule surprisal (based on the description length of the rule used in a rule application, if present), and ``noise'' surprisal (directly related to edit distance).
More formally, for a sequence of Dutch words $\mathbf{y} = y_1, y_2, \dots y_T$ and a sequence of English words $\mathbf{z} = z_1, z_2, \dots z_T$, we plot $p(z_t \mid z_1, z_2, \dots z_{t-1})$ and $p(z_t \mid y_t)$.
When $\beta = 0$, we see that the latent English sentence (the inferred translation) is dominated by the noise likelihood. In other words, it prioritizes keeping string edit distance small with no regard to the plausibility of $z$ as an English sentence, leading to low-probability sequences of English words which have high surface-level similarity to the Dutch sentence. 
As $\beta$ increases, the inference process gradually increases the contribution of the English LM. 
In the figure, this is reflected in particles with reduced LM surprisal, possibly at the cost of increased transformation surprisal. The inferred translation of the Dutch sentence more closely resembles a plausible English sentence, rather than a word-by-word mapping based only on surface-level similarity. In particular, we observe that English words which are quite dissimilar from their Dutch counterparts are inferred, which can be explained by the use of rejuvenation moves which propose word transformations by applying the LM's next-word prediction ability the prefix of an inferred latent English sentence. If the current particle's inferred latent sentence is of low quality, such moves are unlikely to be accepted under Metropolis-Hastings. Yet once the current particle's sentence is close enough to a reasonable English sentence to start making predictions, the quality of these proposals also increases. 
This results in a phase-shift pattern where inference may in some cases undergo rapid movement to a high-probability region of the problem space, while in other cases it remains stuck in low-probability regions.  

\section{Evaluation}

\subsection{Data}

Given that intercomprehension for humans is most likely to succeed when dealing with simple sentences with a high density of cognates, we do not expect our model to translate arbitrary sentences well from an unfamiliar language. However, we do expect to see meaningful variation in the success of the model's zero-shot approach when given sufficiently simple and cognate-rich sentences. 
To this end, we create a basic evaluation dataset consisting of three language pairs: Dutch-English, Italian-Spanish, and Ukrainian-Russian. 
For each language pair, parallel sentences are extracted from the OpenSubtitles corpus and filtered for length and quality. More information on dataset construction is found in \Cref{app:dataset}.

\subsection{Evaluation Metrics}

To compare the similarity of two L1 translations of a given L2 sentence (e.g. human-model similarity or model-ground truth similarity), we employ three distinct metrics: 
(1) Character-level edit distance is the number of insertions, deletions, or substitutions, needed to transform one string into another. (2) Word-level edit distance is analogous, but operates on word edits rather than character edits. 
(3) Sentence embedding cosine distance using SBERT \citep{reimersSentenceBERTSentenceEmbeddings2019} for L1=English or Multilingual E5 \citep{wangMultilingualE5Text2024} for the non-English L1 settings. 
For a particular item, we compare the similarity of the \textit{distribution} of human responses to the \textit{distribution} of model responses using the Wasserstein or Earth-Mover's Distance, as described in \Cref{app:wasserstein}. Intuitively, this captures the divergence between two discrete distributions, using one of the above three distance functions. 

\subsection{Ablations and Baselines}

We perform inference using the following ablations and baselines:

\textbf{Likelihood-only baseline without LM prior tempering.}
This ablation does not perform iterative inference with gradual tempering in of the LM prior. L1 candidates are chosen purely based on likelihood -- i.e., by choosing the L1 word which has the smallest edit distance from the observed L2 word. This minimizes the edit distance between observation and inferred translation, but does not enforce a pressure for plausible L1 messages. 

\textbf{Remove rule library proposals from inference.}
This ablation implicitly enforces a constraint that every word in L2 must be a cognate of some word in L1, and scores mappings based only on their edit distance. This fails to account for the ability to infer and memorize a reusable rule that is based on context and not on any surface-level similarity.  

\textbf{Reduce the number of particles used in SMC inference.} 
This manipulation reflects the inverse relationship between computational resources and exactness of inference in Sequential Monte Carlo. We use 32 particles for the full model and 1 particle for the ablated model. Thus, the ablated model reflects greedy inference which only ever maintains one hypothesis at a time, while SMC with 32 particles reflects parallel processing via a weighted collection of hypotheses. 

\textbf{Zero-Shot Prompted LLMs.} 
As an alternative to our Bayesian inference approach, we attempt zero-shot translation using the following prompted LLMs: \verb|Gemma-3-1B|, \verb|Llama-3.2-1B|, and \verb|SmolLM-2-135M| (for English-Dutch). The prompt used can be found in \Cref{app:llm-baseline-prompt}. The \verb|SmolLM-2-135M| model is parameter-matched to the Goldfish models used by our approach; the Gemma and Llama models are much larger at 1 billion parameters. None of these baseline models are strictly trained on monolingual data, though \verb|SmolLM| only officially supports English. 


\section{Human Experiment}

We perform a behavioral experiment to establish intercomprehension norms for the 100 sentences in each of Dutch, Italian, and Ukrainian; sentences in these languages were presented to native English, Spanish, and Russian speakers, respectively.  
For each language, we recruit 200 participants on Prolific. More details on the human experiment are provided in \Cref{app:human-exp-details}.

\section{Results}

\begin{figure}[htb]
    \centering
    \includegraphics[width=\linewidth]{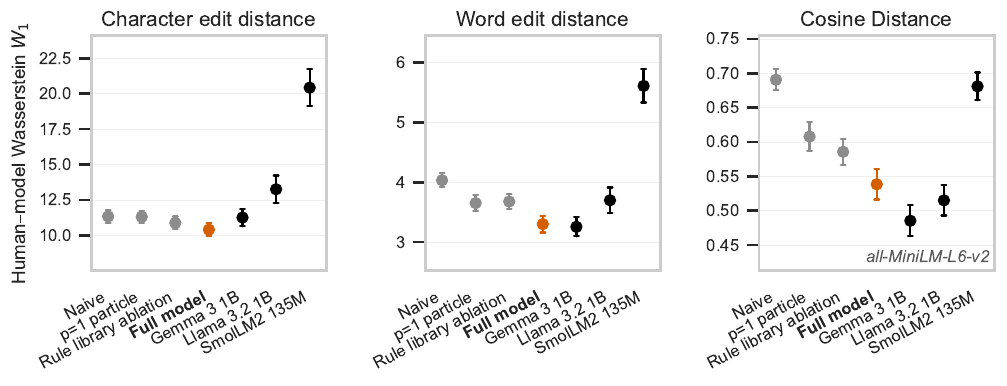}
    \caption{Dutch (L2) to English (L1).}
    \label{fig:model-comparisons}
\end{figure}

\begin{figure}[htb]
    \centering
    \includegraphics[width=\linewidth]{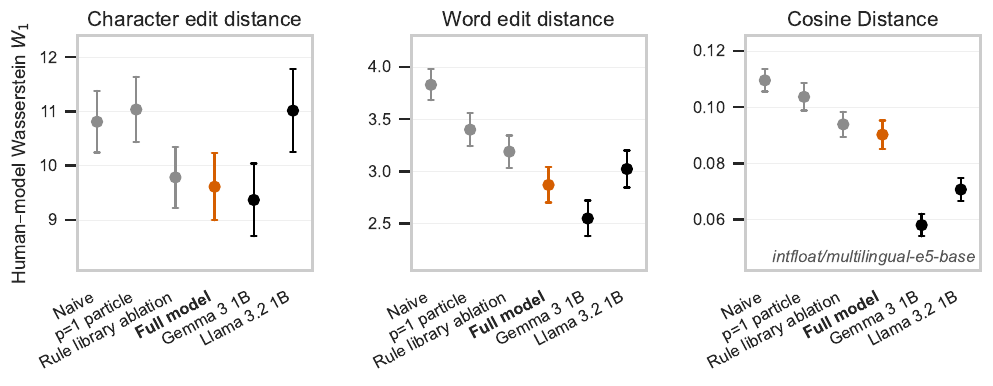}
    \caption{Italian (L2) to Spanish (L1).}
    \label{fig:model-comparison-es-it}
\end{figure}

\begin{figure}[htb]
    \centering
    \includegraphics[width=\linewidth]{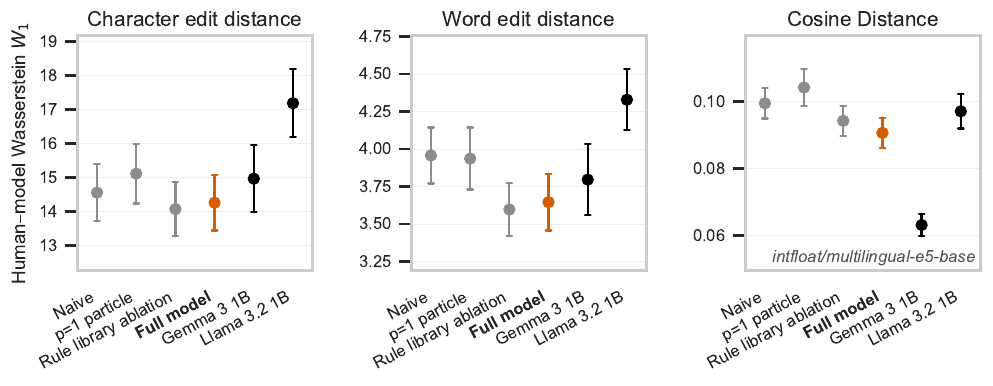}
    \caption{Ukrainian (L2) to Russian (L1).}
    \label{fig:model-comparison-ru-uk}
\end{figure}

\begin{figure}
    \centering
    \includegraphics[width=0.95\linewidth]{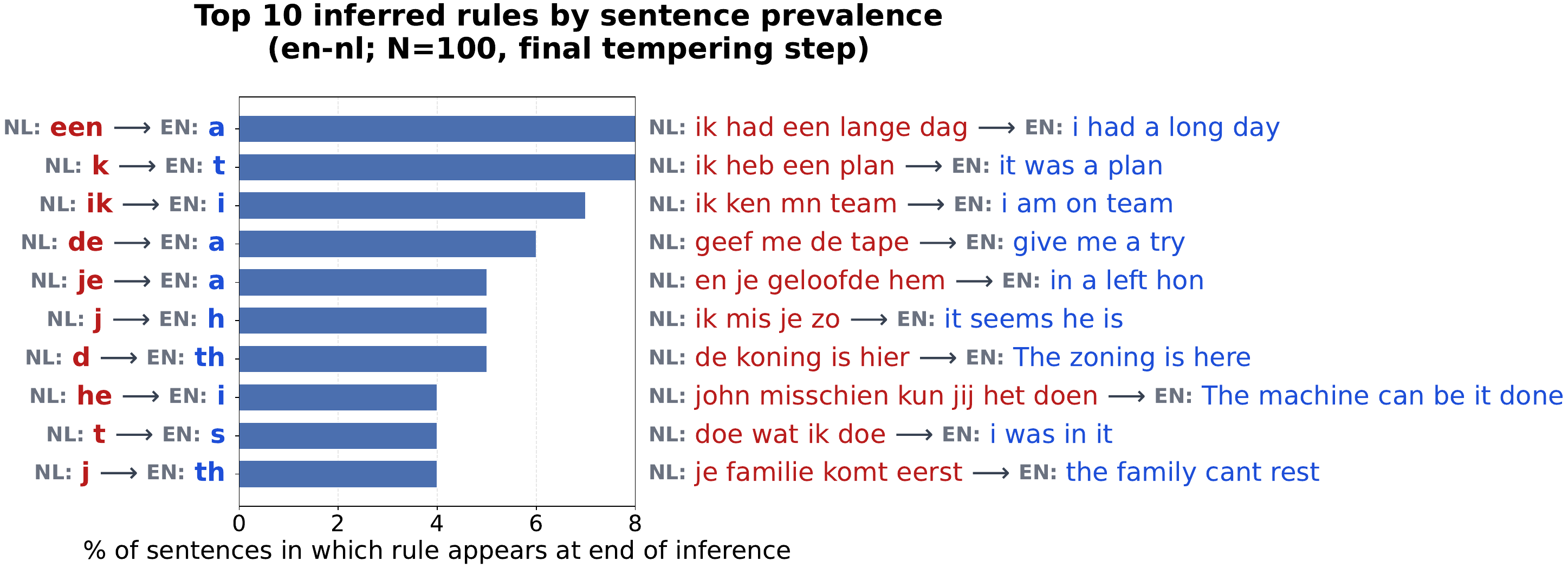}
    \caption{Top inferred subword rules for Dutch $\rightarrow$ English. The patterns show both correct inferences (\textit{een} $\rightarrow$ \textit{a}), as well as incorrect inferences (\textit{k} $\rightarrow$ \textit{t}).} 
    \label{fig:inferred-rules-en-nl}
\end{figure}

\subsection{Model demonstrates graded success at intercomprehension task}

A comparison of the distribution of evaluation scores in humans and models, across evaluation metrics and model settings, is shown in \Cref{fig:model-comparisons,fig:model-comparison-es-it,fig:model-comparison-ru-uk}. We note that the progression of distance metrics goes from most low-level (character edit distance) to most high-level (SBERT embedding cosine distance). 
Several general patterns emerge. 
The particle ablation illustrate the contribution of parallel processing (multiple particles > single particle) 
Likewise, the rule library ablation demonstrates that the ability to infer symbolic rewrite rules leads to greater alignment of model inferences with human inferences. 
Comparing the Naive model to the Full model, we observe that the Full model (which reflects a gradually tempered L1 language model prior) consistently outperforms the Naive model (which uses only form-based similarity) at matching human inferences. 
Under the Tempered SMC algorithm, increasing step number during inference corresponds to an increasing contribution of the L1 language model prior. The results thus demonstrate that gradually introducing L1 priors leads to improvement over a naive, exclusively surface-form-based model of inference. 

\subsection{Inferred rules capture both cognate structure and red herrings}

A direct byproduct of SMC inference with the provided generative model is a distribution over inferred rules. We visualize the most commonly inferred rules in \Cref{fig:inferred-rules-en-nl}. For each Dutch$\rightarrow$English re-write rule, one randomly chosen exemplar item is shown (a Dutch sentence in the evaluation corpus, and the inferred English sentence formed using that rule). 
Full-word rewrite rules like \textit{een}$\rightarrow$\textit{a} and \textit{ik}$\rightarrow$\textit{i} are common. The subword rule \textit{d}$\rightarrow$\textit{th} reflects the correct cognate correspondence between English \textit{th} and Dutch \textit{d}.
As would be expected, however, there are also incorrect inferred rules, or ``red herrings''. Because the model is operating in a very low-data setting (single sentence observations), and has no prior knowledge of cognate relationships between languages, rules may be inferred that results in plausible-seeming inferences, even if they are not veridical.

Top inferred rules for the Italian-Spanish and Ukrainian-Russian language pairs are shown in \Cref{app:rules-analysis-nonenglish}.

\subsection{Comparison to zero-shot prompted LLMs}

In many cases, our approach achieves comparable or even closer alignment to human inferences, compared with much larger prompted LLMs, despite these larger models not being trained on strictly monolingual data. In the English-Dutch experiments, we performed zero-shot prompting with the parameter-matched \verb|SmolLM| model, which aligns with human inferences substantially worse on every metric. 
Our approach is successful specifically because it decomposes intercomprehension into language modeling and symbolic inference, and only uses the small Goldfish model for what they are trained to do: capture the prior over messages in L1. 
Our model tends to be outperformed by Gemma and Llama when judged by the most high-level, semantic metric (SBERT). This suggests that our model's word-for-word translation approach fails to capture some aspects of human intercomprehension, which may be better modeled using gist-based approaches. 



\section{Discussion}

\subsection{Cognitively plausible model of intercomprehension}

The model described in this paper is intended as an algorithmic-level model of a cognitive phenomenon, not as a state-of-the-art model for machine translation. Clearly, higher performance on translation can be achieved by training models on large multilingual (and possibly parallel) datasets. Yet the problem of zero-shot decipherment of an unfamiliar language, using only knowledge of a related language and general-purpose inference strategies, reflects a cognitive task with ecological validity and which requires flexible, real-time probabilistic inference. 
Our key innovation is to extend psycholinguistic models of noisy-channel inference \citep{levyNoisyChannelModelHuman2008, gibsonRationalIntegrationNoisy2013, clarkModelApproximateIncremental2025}, typically applied to out-of-distribution utterances \textit{within} a language, to a much more general case of inference across potentially \textit{different} languages. 
Key to achieving this is an efficient approximate inference algorithm that incorporates a cognitively plausible generative model of language transformations, and heuristics for performing custom inference within the SMC framework. The result is a model which requires no exposure to Dutch, Dutch-specific heuristics, or training on parallel corpora in order to achieve human-like intercomprehension ability. 

The model outlined in this paper instantiates resource-rational inference \citep{liederResourcerationalAnalysisUnderstanding2020}, where the number of SMC particles is a loose proxy for cognitive resources used to process a sentence. 
Unlike past work which used SMC to model incremental language comprehension in terms of sequential, word-by-word processing \citep{levyModelingEffectsMemory2008, clarkResourceRationalNoisyChannelLanguage2025}, our model applies SMC to iterative subproblems in a different sense: by gradually tempering in the contribution of an L1 prior, we model a comprehender who moves from inferences about surface-level similarity towards richer inferences that integrate priors about intended messages. 
Crucially, the ability to ground inference in similarity as a starting point allows successive improvements that unfold over time, modeling intercomprehension not as strictly left-to-right reading, but as an iterative refining and sharpening of inferences in a non-sequential way. 
A computational model of intercomprehension provides a useful tool for testing hypotheses about the strategies and algorithms used by human comprehenders (which can be added or ablated from the current model), and for explaining phenomena such as asymmetric intelligibility (i.e. cases where speakers of Language A can understand Language B better than speakers of Language B can understand Language A). 

\subsection{Parallel processing and symbolic rules as core model components}

Our ablations analysis demonstrates that two specific model features contribute to improved scores on the evaluation metrics.
First, the use of parallel processing with 32 particles in an SMC inference framework produced improved scores relative to single-particle inference. A collection of multiple weighted particles appears better able to represent the wide uncertainty about the possible translation equivalents of a given word sequence during iterated inference, whereas a single particle may greedily commit to one option out of many and fail to more fully explore the space of alternatives.  
Second, the ability to infer word re-write rules enables a ``stab in the dark'' approach to translation; ablated runs without this feature, which were limited to proposing latent English strings using only surface form similarity, performed worse.
When only a few words in a sentence remain to be translated, humans can often make an informed guess about the identities of the remaining words using context as a guide, e.g. as seen in the cloze task \citep{taylorClozeProcedureNew1953, staubInfluenceClozeProbability2015, kuperbergWhatWeMean2016a}. If the guess results in a significantly better translation, it might be accepted even if the inferred L1 word bears little resemblance to the observed L2 word. 
This type of ``a-ha moment'' or sudden insight is studied in other contexts in cognitive science \citep{goodmanRationalAnalysisRuleBased2008, tenenbaumTheorybasedBayesianModels2006, kouniosCognitiveNeuroscienceInsight2014, ruleLearningListConcepts2018,dubeyAhaMomentsCorrespond2025}, and here we provide an algorithmic correlate of this type of inference under high uncertainty, which can nevertheless result in a successful belief update.


\subsection{Augmenting small LLMs with Bayesian inference}

Thanks to their pre-training, modern LLMs are strong multi-task learners \citep{radfordLanguageModelsAre2019}. 
However, in cases where a particular problem admits decomposition into language modeling and symbolic structure, the performance of small LLMs can be improved using Bayesian inference with explicit, customizable generative models. This is consistent with past work that has leveraged SMC inference to enforce explicit constraints on language models \citep{loulaSyntacticSemanticControl2025, lipkinFastControlledGeneration2025}. In domains with strong constraints or structural priors, or where the inputs are highly out-of-distribution relative to the training data, or where model size is a large constraint, approaches like the one employed in this work may have broader applicability beyond the cognitive modeling of intercomprehension. 

\subsection{Limitations and Future Work}

The current work only considers intercomprehension in a reading setting, rather than a listening setting. Reading is different from listening in that orthographic cues play a large role, while phonetic cues are mediated by a reader's ability to infer how an unfamiliar word should be pronounced. 
Another major limitation of the generative model in its current form is the strong assumption of word order alignment between L1 and L2 sentences. Future iterations of this model can consider additional word order transformations as part of the generative model (e.g. a mapping from the positions of L1 words to the positions of their L2 equivalents), and a corresponding proposal function for inference which proposes changes to a particle's latent word order mapping. This approach would still face challenges in dealing with $L1 \rightarrow L2$ mappings that are not one-to-one, e.g. ``not'' (English) $\rightarrow$ ``ne ... pas'' (French). Additionally, allowing changes to word order between L1 and L2 sentences results in a combinatorial explosion in the number of transformations of a particular sentence, making inference significantly more difficult.
However, the language pairs which have high rates of mutual intelligibility tend to be those which have similar word order rules \citep{gooskensMutualIntelligibilityClosely2018}, so this limitation may have a cognitive interpretation.






\bibliography{colm2026_conference}
\bibliographystyle{colm2026_conference}

\appendix
\crefalias{section}{appendix}

\clearpage

\section{Full Generative Model Specification}
\label{app:model-spec}

The generative model defines a joint distribution over an $L_1$ sentence $S$, and a rule library $\mathcal{L}$ given an observed $L_2$ sentence $\mathbf{y} = (y_1, \ldots, y_T)$.

\paragraph{$L_1$ sentence prior.}
\[
S \sim p_{\mathrm{LM}}(\cdot)
\]
where $p_{\mathrm{LM}}$ is a Goldfish language model distribution over L1 word sequences. We use the following Goldfish models:

\begin{table}[h]
\centering
\renewcommand{\arraystretch}{1.5}
\begin{tabular}{llcc}
\hline
\textbf{L1} & \textbf{Model String} & \textbf{Training Data Size} & \textbf{\# Parameters} \\
\hline
English & \verb|goldfish-models/eng_latn_1000mb| & 1000 MB & 125M \\
Spanish & \verb|goldfish-models/spa_latn_1000mb| & 1000 MB & 125M \\
Russian & \verb|goldfish-models/rus_cyrl_1000mb| & 1000 MB & 125M \\
\hline
\end{tabular}
\caption{Caption}
\label{tab:my_table}
\end{table}

Sentence probability is additionally weighted by $p_{\mathrm{EOS}}(S)$, the LM probability of an end-of-sentence token following $S$, which favors grammatically complete sentences.

\paragraph{Rule library prior.}
A rule library $\mathcal{L}$ is a set of string rewrite rules $r = (r_{\mathrm{in}},\, r_{\mathrm{out}})$, where $r_{\mathrm{in}}$ is a regex pattern and $r_{\mathrm{out}}$ a replacement string. The prior penalizes complex libraries:
\[
\log p(\mathcal{L}) = -\lambda_{\mathcal{L}} \sum_{r\,\in\,\mathcal{L}} \bigl(|r_{\mathrm{in}}| + |r_{\mathrm{out}}|\bigr)
\]
where $|\cdot|$ is character length and $\lambda_{\mathcal{L}} = 2.0$.

\paragraph{Per-word noise model.}
Let $w_t$ be the $t$-th word of $S$, and let
\[
\mathcal{L}(w_t) = \bigl\{ r(w_t) : r \in \mathcal{L},\; r_{\mathrm{in}} \text{ matches } w_t \bigr\}
\]
be the set of words obtainable by applying a matching library rule to $w_t$. For each position $t = 1,\ldots,T$:
\[
\mathrm{applyRule}_t \sim \operatorname{Bernoulli}\!\left(
\begin{cases}
0.9 & \text{if } \mathcal{L}(w_t) \neq \emptyset \\
0   & \text{otherwise}
\end{cases}
\right)
\]
\[
\log p\!\left(y_t \mid w_t,\, \mathcal{L},\, \mathrm{applyRule}_t\right) =
\begin{cases}
\log \dfrac{1}{|\mathcal{L}(w_t)|} \cdot \mathbf{1}\!\left[y_t \in \mathcal{L}(w_t)\right]
  & \text{if } \mathrm{applyRule}_t = 1 \\[8pt]
-\lambda_{\mathrm{noise}} \cdot \mathrm{EditDist}(w_t,\, y_t)
  & \text{if } \mathrm{applyRule}_t = 0
\end{cases}
\]
where $\lambda_{\mathrm{noise}} = 2.0$ and $\mathrm{EditDist}$ is the simple mean of the character-level edit distance (orthographic likelihood) or ARPAbet phoneme edit distance based on a nonce-word pronunciation module (phonetic likelihood).

\paragraph{Joint distribution.}
\begin{align*}
p\!\left(S, \mathcal{L}, \{\mathrm{applyRule}_t\} \mid \mathbf{y}\right) \;\propto\;
  &\; p_{\mathrm{LM}}(S)\; p_{\mathrm{EOS}}(S)\; p(\mathcal{L}) \\
  &\;\times \prod_{t=1}^{T} p\!\left(\mathrm{applyRule}_t \mid w_t, \mathcal{L}\right)
              p\!\left(y_t \mid w_t, \mathcal{L}, \mathrm{applyRule}_t\right)
\end{align*}

\clearpage

\section{Inference details}
\label{app:inference-details}

\paragraph{Tempered SMC inference target.}
Inference uses tempered SMC with a linear inverse-temperature schedule $\beta_k = k/K$, $k = 0,\ldots,K$. At step $k$ the target is:
\[
p_{\beta_k}\!\left(S, \mathcal{L} \mid \mathbf{y}\right) \propto
  p_{\mathrm{LM}}(S)^{\beta_k}\; p(\mathcal{L})
  \prod_{t=1}^{T} p\!\left(y_t \mid w_t, \mathcal{L}\right)
\]
At $\beta_0 = 0$ the target is determined solely by the edit-distance likelihood; at $\beta_K = 1$ it equals the full posterior. 

\paragraph{Weight updates.}
For a sequence of $\beta_k$ values $0 < \beta_1 < \beta_2 < \dots < \beta_K = 1$, we compute an incremental importance ratio, which bridges between one target distribution and the next:
$$\frac{\pi_{\beta_k}(z)}{\pi_{\beta_{k-1}}(z)} \propto \frac{p(y \mid z) p(z)^{\beta_k}}{p(y \mid z) p(z)^{\beta_{k-1}}} = p(z)^{\Delta\beta_k}$$
Thus the weight update for each particle is given by: $w_k^{(i)} \leftarrow w_{k-1}^{(i)} \cdot p \left (z^{(i)} \right )^{\Delta\beta_k}$.
At the final step of inference, the algorithm yields a weighted collection of particles corresponding to approximate samples from the final posterior target distribution $\pi_{\text{post}}(z)$.  

\paragraph{Resampling.}
After each weight update, we compute the Effective Sample Size (ESS) using the formula $\text{ESS} = ( \sum w^{(i)} )^2 / \sum ( w^{(i)} )^2$. We resample the particles if the ESS drops below a preset value, using systematic resampling to reduce variance \citep{carpenterImprovedParticleFilter1999}. 

\paragraph{Metropolis-Hastings Accept-Reject Moves.}
Each proposal function defines a transition probability $q(z^\prime \mid z)$ from one particle latent state to another. 
Using the Metropolis-Hastings algorithm, the acceptance probability for a rejuvenation move is given by:
$$\alpha = \min \left ( 1, \frac{p(y \mid z^\prime) \: p(z^\prime)^{\beta_k} \: q(z \mid z^\prime)}{p(y \mid z) \: p(z)^{\beta_k} \: q(z^\prime \mid z)} \right )$$
This quantity is tracked automatically given the fully specified generative model using the Gen probabilistic programming language \citep{cusumano-townerGenGeneralpurposeProbabilistic2019, cusumano-townerAutomatingInvolutiveMCMC2020}, which includes Metropolis-Hastings as part of the standard inference toolkit. As a practical note, since the inference process requires many calls to the language model, often with identical arguments across iterations and particles, significant efficiency gains result from utilizing a cached LM, following \citet{lewSequentialMonteCarlo2023}.
We design kernels to have the property that the quantities $q(z \mid z^\prime)$ and $q(z^\prime \mid z)$ are both non-zero: in other words, we enforce the property that each kernel which updates a particle $x$ to a proposed new particle $x^\prime$ be capable of inverting this update by calling itself on the new particle with some probability.

\clearpage

\section{Datasets}
\label{app:dataset}

For each L1-L2 language pair, the evaluation dataset is drawn from the OpenSubtitles parallel movie subtitles dataset accessed via OPUS \citep{tiedemannOPUSMTBuildingOpen2020} by randomly sampling 100,000 sentences. From this starting sample, we filter out sentences longer than 10 words or shorter than 4 words; where the L1 and L2 versions of the sentence differ in the number of words; which contain special symbols or non-standard punctuation; which contain capitalized words not at the beginning of the sentence (to remove proper nouns and acronyms); where the L1 translation contains words not in a wordlist of the top $V$ L1 words according to the \verb|wordfreq| package.
We then apply dependency parsing with spaCy \citep{honnibalSpaCyIndustrialstrengthNatural2020} to sentences (both L1 and L2), classify sentence pairs as having matching or non-matching dependency structures, and sample 150 sentences from each category to introduce variation in sentence structure match. 
Next, we apply an independent automated machine translation step to the L2 sentences using \verb|meta-llama/Llama-3.3-70B-Instruct|.  
We filter out sentences where the machine-translated L1 output does not match the ground truth L1 sentence in OPUS. 
This filters out non-literal, idiomatic translations, e.g. \textit{M'n baard is korter} (Dutch), which is translated as ``I shortened my beard'' in the parallel English subtitles but as ``My beard is shorter'' by machine translation (a more literal translation). 
As a final step, filtered sentences are manually reviewed to exclude potentially offensive content. 
The result of this process is a set of 100 parallel simple sentences, which are selected to elicit a dynamic range of intercomprehension from L1 speakers who are otherwise unfamiliar with the L2 (we note that these are not unbiased samples from the baseline distribution over L2 sentences).

\section{Computation of the Human--Model Wasserstein Distance}
\label{app:wasserstein}

For each item we compare two distributions over sentences: the empirical distribution of human responses and the model's posterior over candidate translations. We summarize their discrepancy with the discrete $1$-Wasserstein distance (earth mover's distance) under three sentence-level ground metrics. The distance is computed exactly per item by solving the optimal-transport linear program; we do \emph{not} use the closed-form one-dimensional Wasserstein estimator.

\paragraph{Distributions.}
Let an item have $n$ human responses $h_1,\dots,h_n$. We place the uniform empirical measure $a = (1/n,\dots,1/n)$ over them, so identical responses contribute proportionally to their frequency. For model inferences, the log particle weights $w_p$ are exponentiated and normalized, $\tilde{w}_p = e^{w_p}/\sum_{p'} e^{w_{p'}}$, each particle's final sentence is standardized (lower-cased, stripped of punctuation), and the weights of identical standardized sentences are summed. 
This yields a distribution $b = (b_1,\dots,b_m)$ over $m$ distinct model sentences $g_1,\dots,g_m$, with $\sum_k b_k = 1$.

\paragraph{Ground metrics.}
For each pair $(h_i, g_k)$ we form a cost $C_{ik}$ under the three metrics of character edit distance, word edit distance, and sentence embedding cosine distance, producing three cost matrices $C \in \mathbb{R}^{n \times m}$.

\paragraph{Optimal transport.}
The $1$-Wasserstein distance is the minimum-cost coupling between $a$ and $b$:
\begin{equation}
  W_1(a, b; C) \;=\; \min_{\gamma \in \mathbb{R}_{\ge 0}^{n \times m}} \;\sum_{i=1}^{n} \sum_{k=1}^{m} \gamma_{ik}\, C_{ik} \quad \text{s.t.} \quad \sum_{k} \gamma_{ik} = a_i, \;\; \sum_{i} \gamma_{ik} = b_k .
\end{equation}
We solve this linear program exactly with the HiGHS solver (\texttt{scipy.optimize.linprog}), treating $\gamma$ as $nm$ nonnegative variables with objective vector $\mathrm{vec}(C)$. The constraint set comprises the $m$ column-marginal equalities and $n-1$ of the row-marginal equalities (one row constraint is omitted as redundant, since both marginals are normalized). The optimal objective value is reported as $W_1$; the recovered $\gamma$ gives the optimal transport plan. We compute $W_1$ separately for each of the three cost matrices, yielding three values per item.

\clearpage

\section{Human Experiment Details}
\label{app:human-exp-details}

\begin{figure}[htb]
    \centering
    \includegraphics[width=0.6\linewidth]{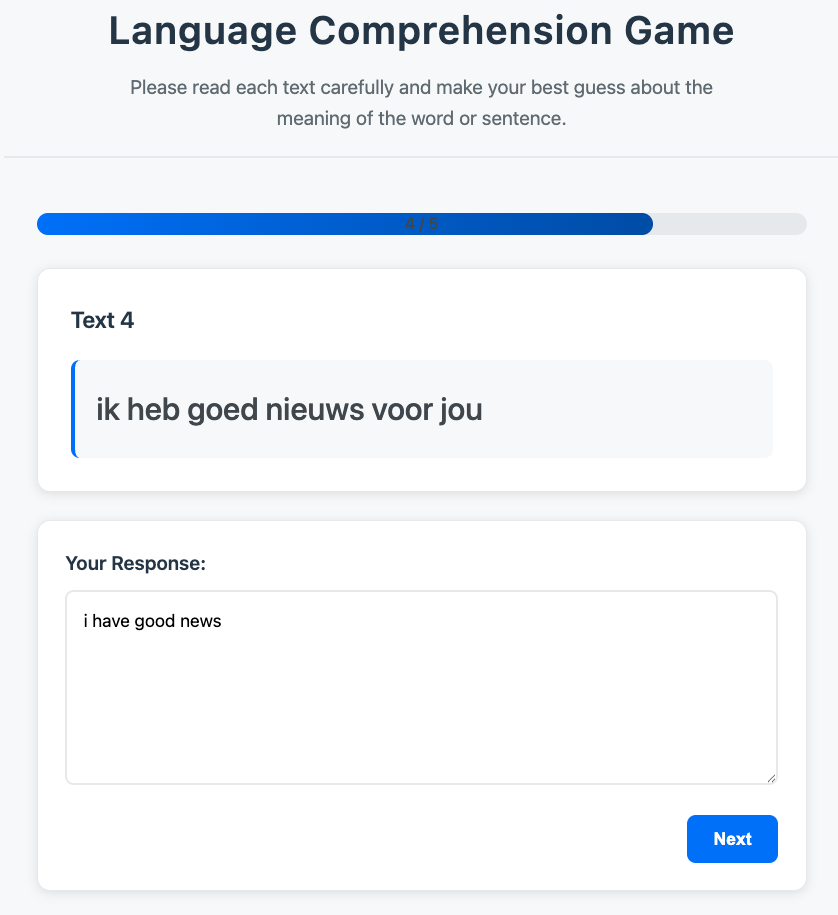}
    \caption{Experimental paradigm.}
    \label{fig:exp-screen}
\end{figure}

Dutch sentences were presented to self-reported English-speaking monolinguals. 
Italian sentences were presented to Spanish speakers; those who reported prior knowledge of Italian were filtered out. 
Ukrainian sentences were presented to Russian speakers; those who reported prior knowledge of Ukrainian were filtered out.
The sentences are divided into 20 disjoint lists, each consisting of 5 unique items. Participants are assigned to experimental lists pseudo-randomly while ensuring balanced numbers of participants across lists. 
All participants provide informed consent at the beginning of the study. 
The study was conducted according to an approved IRB protocol at the authors' institution. 
Participants each saw 5 experimental items. In total, the expected duration of the study was 5 minutes, and participants were paid \$1.25. 
Participants were instructed: ``Please read each text carefully and make your best guess about the meaning of the sentence''.
Participants were explicitly told not to use any artificial intelligence (AI) tools, translation apps, or other outside help during the task. Additional steps were taken to prevent use of outside tools: we blocked the ability to select and copy or paste text. 

\Cref{fig:exp-screen} shows a screenshot of the experimental interface for the human behavioral experiment. 

\clearpage

\section{Example Model Inferences}
\label{app:example-model-inferences}

\begin{center}
\scriptsize
\begin{longtable}{p{\dimexpr 0.27\linewidth-2\tabcolsep\relax}%
                  p{\dimexpr 0.365\linewidth-2\tabcolsep\relax}%
                  p{\dimexpr 0.365\linewidth-2\tabcolsep\relax}}
\toprule
\textbf{Dutch Sentence} & \textbf{Human Inferences} & \textbf{Model Inferences} \\
\midrule
\endfirsthead
\multicolumn{3}{c}{\tablename~\thetable{} (continued)} \\
\toprule
\textbf{Dutch Sentence} & \textbf{Human Inferences} & \textbf{Model Inferences} \\
\midrule
\endhead
\midrule
\multicolumn{3}{r}{\textit{Continued on next page}} \\
\endfoot
\endlastfoot
\textit{slaap is het beste} \newline (sleep is the best) & slap is the best\newline sleep is the best & sleep is a beast\newline sleep is the best \\\midrule
\textit{geef haar een magisch moment} \newline (give her a magical moment) & Give her a magical moment\newline Good have a magic moment & geek hair and magic moments\newline gee hear a magic moment \\\midrule
\textit{het is een vorm van communicatie} \newline (it is a form of communication) & He is bad at communicating\newline what is in worm can communicate & It is a form of communication\newline hat is a form of communication \\\midrule
\textit{wij hebben de papieren} \newline (we have the papers) & {Do we have any paper}\newline{Give me the paper} & we happen on paper\newline we happen to be \\\midrule
\textit{het was een droom} \newline (it was a dream) & He was in a room\newline He was in his room & it was a dream\newline i was in doom \\\midrule
\textit{ik heb goed nieuws voor jou} \newline (i have good news for you) & I have good news for you\newline I heard good news for you & i have good news for you\newline we have good news for you \\\midrule
\textit{hij werkt te hard} \newline (he worked too hard) & He worked too hard \newline He's worked too hard & the work is hard\newline i worked it hard \\\midrule
\textit{we hadden kalm moeten blijven}\newline (we should have stayed calm) & We had a calm moderate blizzard \newline We had a calm modest evening & the sudden calm met oblivion \newline a sudden calm met oblivion \\\midrule
\textit{dit is een militaire post}\newline (this is a military post) & {Did it happen at the military post?}\newline{It is a military base} & this is a military post\newline it is a military post \\
\bottomrule
\noalign{\vspace{\abovecaptionskip}}
\captionsetup{font=normalsize}
\caption{Example human and model inferences for Dutch sentences from the dataset. Each sentence is shown with two human-generated and two model-generated inferences.}
\label{tab:example-inferences}
\end{longtable}
\end{center}

\clearpage

\section{Inferred rules analysis for non-English language pairs}
\label{app:rules-analysis-nonenglish}

\begin{figure}[htb]
    \includegraphics[width=0.95\linewidth]{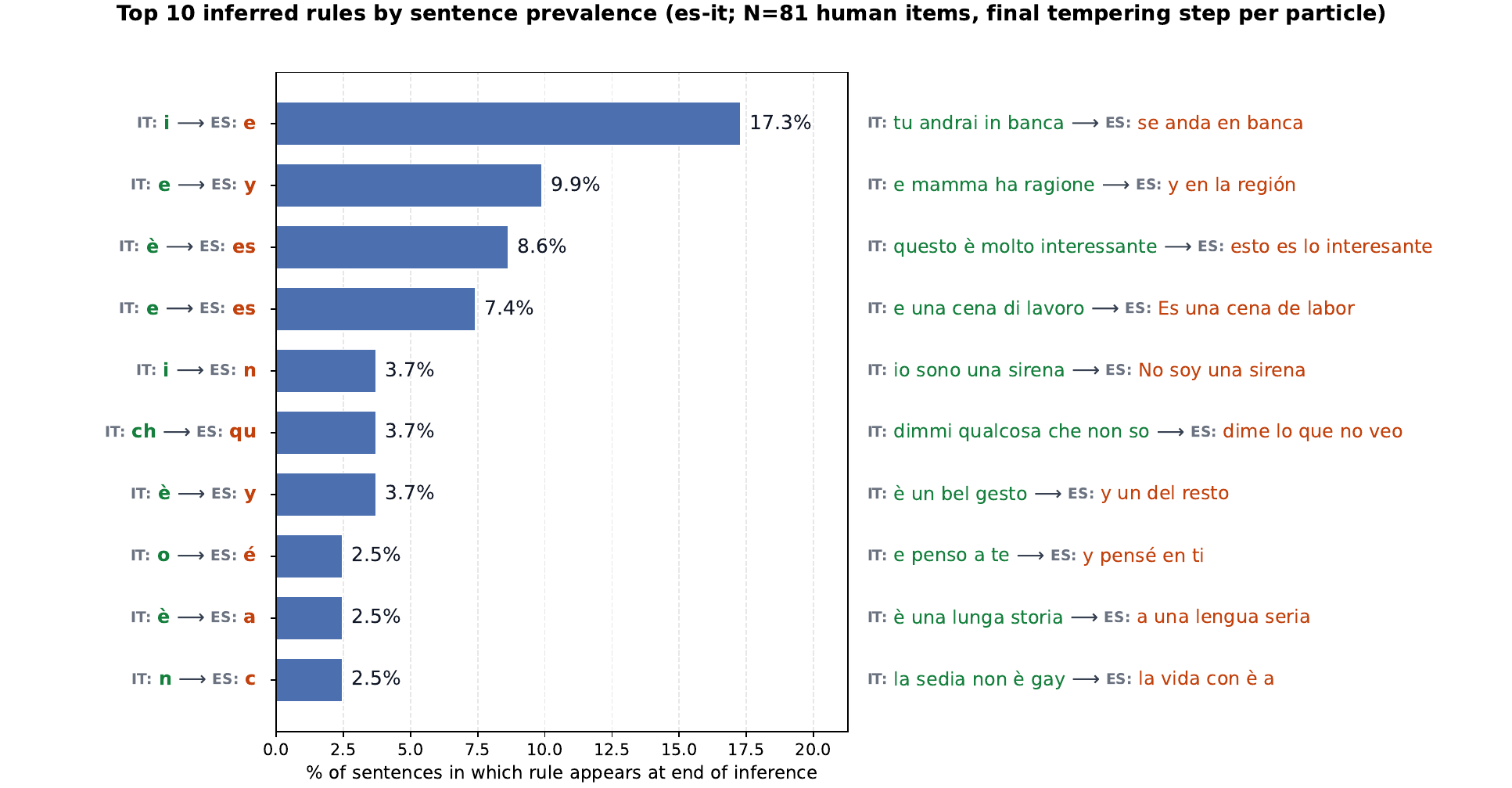}
    \caption{Top inferred subword rules for Italian $\rightarrow$ Spanish. The patterns show both correct inferences (\textit{ch} $\rightarrow$ \textit{qu}), as well as incorrect inferences (\textit{i} $\rightarrow$ \textit{n}).}
    \label{fig:inferred-rules-es-it}
\end{figure}

\begin{figure}[htb]
    \includegraphics[width=0.95\linewidth]{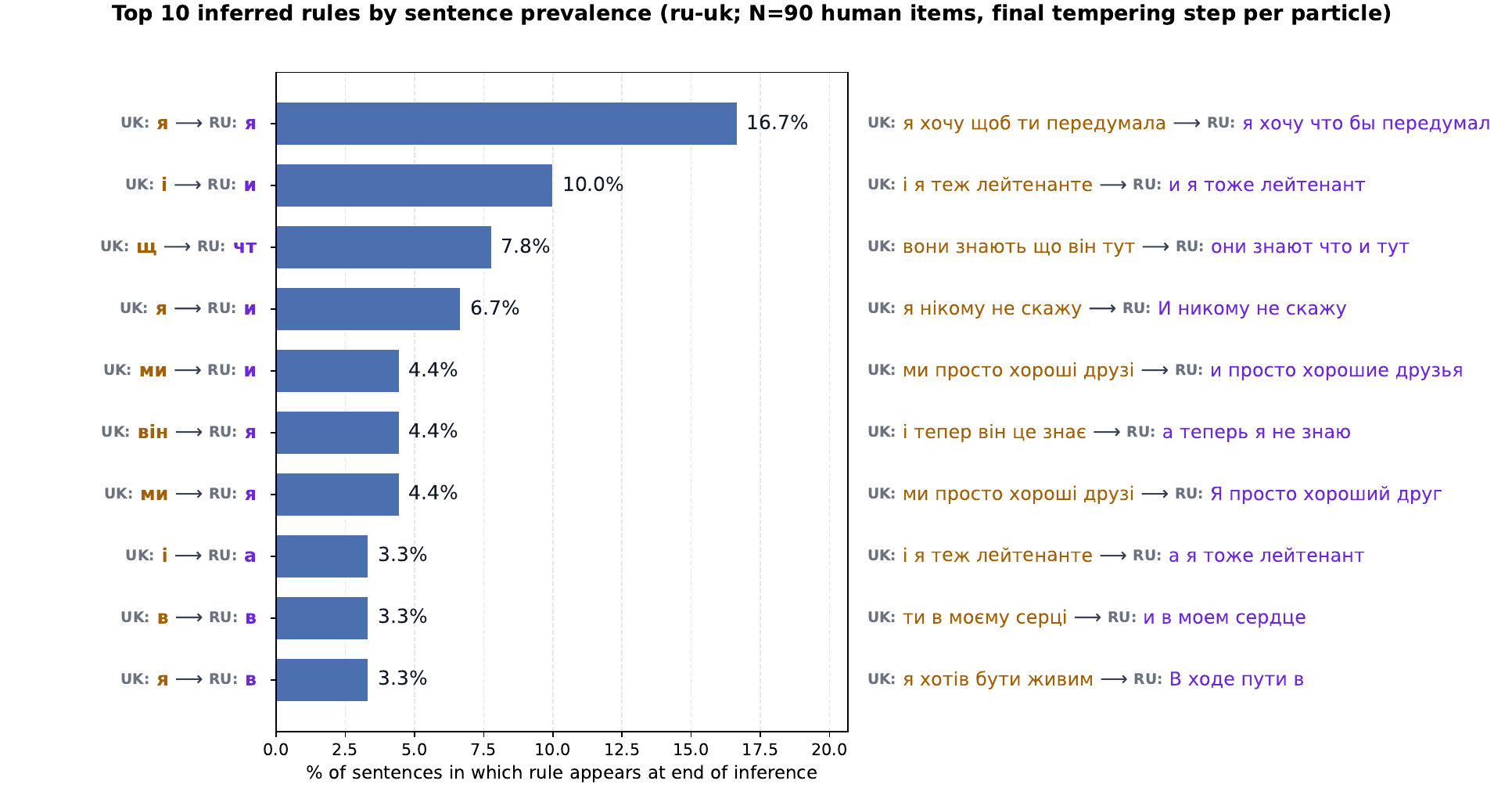}
    \caption{Top inferred subword rules for Ukrainian $\rightarrow$ Russian. The patterns show both correct inferences (\foreignlanguage{russian}{щ} $\rightarrow$ \foreignlanguage{russian}{чт}), as well as incorrect inferences (\foreignlanguage{russian}{він} $\rightarrow$ \foreignlanguage{russian}{я}).}
    \label{fig:inferred-rules-ru-uk}
\end{figure}

\clearpage

\providecommand{\cyr}[1]{{\fontencoding{T2A}\selectfont #1}}

\section{Zero-Shot Prompting of the LLM Baselines}
\label{app:llm-baseline-prompt}

The LLM baselines (Llama 3.2 1B, Gemma 3 1B, and SmolLM2 135M) are prompted zero-shot: each stimulus sentence is sent as a single user turn to the model's chat endpoint, with no system prompt, no in-context examples, and no conversation history. The prompt names only the target (reference) language and instructs the model to emit the translation alone; it does not name the source (stimulus) language, so the model must recognize the unfamiliar language on its own, enabling more fair comparison to our model. Each item is translated independently in a fresh context.

\begin{examplebox}{Zero-Shot Translation Prompt (Template)}
> \texttt{Translate \textquotesingle{}\{sentence\}\textquotesingle{} into \{reference\_name\}. Only produce the \{reference\_name\} translation and no other output.}
\end{examplebox}

Here \texttt{\{sentence\}} is the stimulus sentence verbatim and \texttt{\{reference\_name\}} is the human-readable name of the reference language (\emph{English}, \emph{Spanish}, or \emph{Russian}). Instantiating the template for one item from each of the three language pairs gives:

\begin{examplebox}{Zero-Shot Translation Prompt (Instantiated Examples)}
\textbf{Dutch $\rightarrow$ English} \\
> \texttt{Translate \textquotesingle{}calvin had een idee\textquotesingle{} into English. Only produce the English translation and no other output.} \\[4pt]
\textbf{Italian $\rightarrow$ Spanish} \\
> \texttt{Translate \textquotesingle{}abbiamo gi\`a abbastanza problemi\textquotesingle{} into Spanish. Only produce the Spanish translation and no other output.} \\[4pt]
\textbf{Ukrainian $\rightarrow$ Russian} \\
> \texttt{Translate \textquotesingle{}\cyr{а я так і не отримала чек}\textquotesingle{} into Russian. Only produce the Russian translation and no other output.}
\end{examplebox}

\paragraph{Decoding and serving.}
All three baselines are served locally through an OpenAI-compatible chat-completions endpoint in LM Studio\footnote{Version 0.4.16, \url{https://lmstudio.ai/}}. We decode greedily, with temperature $0$, nucleus sampling parameter top-$p = 0.9$, and a cap of $512$ generated tokens; the model's response is stripped of surrounding whitespace and used as the translation with no further post-processing. The specific checkpoints are \texttt{Llama-3.2-1B-Instruct}, \texttt{gemma-3-1b-it}, and \texttt{SmolLM2-135M-Instruct} (bf16 weights).







\end{document}